\documentclass[10pt,twocolumn,letterpaper]{article}

\usepackage{cvpr}              %
\usepackage{graphicx}
\usepackage{amsmath}
\usepackage{amssymb}
\usepackage{booktabs}

\usepackage[pagebackref,breaklinks,colorlinks]{hyperref}

\usepackage[capitalize]{cleveref}
\crefname{section}{Sec.}{Secs.}
\Crefname{section}{Section}{Sections}
\Crefname{table}{Table}{Tables}
\crefname{table}{Tab.}{Tabs.}

\usepackage{url}            %
\usepackage{amsfonts}       %
\usepackage{nicefrac}       %
\usepackage{microtype}      %
\usepackage[dvipsnames]{xcolor}         %
\usepackage[accsupp]{axessibility}
\usepackage{gensymb}
\usepackage{amsmath}
\usepackage[capitalize]{cleveref}
\usepackage{graphicx}
\usepackage{xspace}
\usepackage{multirow}
\usepackage{makecell}

\newcommand{\na}{-}

\newcommand{\objint}{object intrinsics\@\xspace}

\renewcommand{\paragraph}[1]{\vspace{0.1cm}\noindent\textbf{#1}}

\def\cvprPaperID{425} %
\def\confName{CVPR}
\def\confYear{2023}

\usepackage[strict]{changepage}

\usepackage{framed}

\definecolor{formalbar}{rgb}{0.290,0.325,0.337}
\definecolor{formalshade}{rgb}{0.925,0.941,0.976}

\newenvironment{formal}{%
  \MakeFramed{\advance\hsize-\width\FrameRestore}%
  \noindent\hspace{-4.55pt}%
  \begin{adjustwidth}{}{7pt}%
  \vspace{2pt}\vspace{2pt}%
}
{%
  \vspace{-5pt}\end{adjustwidth}\endMakeFramed%
}

\begin{document}

\title{Seeing a Rose in Five Thousand Ways}

\author{Yunzhi Zhang\\
Stanford University\\
\and
Shangzhe Wu\\
University of Oxford\\
\and
Noah Snavely\\
Cornell Tech\\
\and
Jiajun Wu\\
Stanford University
}

\twocolumn[{
\renewcommand\twocolumn[1][]{#1}
\maketitle
\centering
\vspace{-1.9em}
\includegraphics[width=\linewidth]{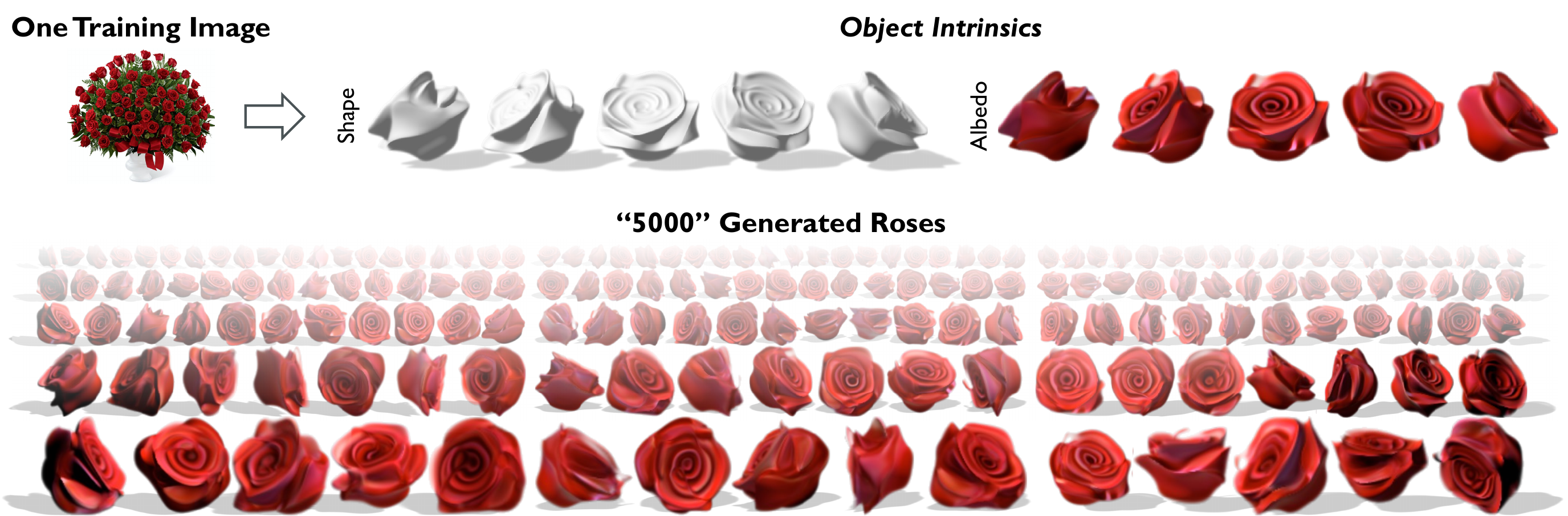}
\vspace{-1.4em}
\captionof{figure}{
From a single image, our model learns to infer object intrinsics---the distributions of the geometry, texture, and material of object instances within the image. 
The model can then generate new instances of the object type, and it allows us to view the object under different poses and lighting conditions.
Project page at {\footnotesize\url{https://cs.stanford.edu/~yzzhang/projects/rose/}}.\protect\footnotemark  
\label{fig:teaser}
}
\bigbreak

}]

\begin{abstract}
\vspace{-5pt}

What is a rose, visually? 
A rose comprises its intrinsics, including the distribution of geometry, texture, and material specific to its object category. 
With knowledge of these intrinsic properties, we may render roses of different sizes and shapes, in different poses, and under different lighting conditions. In this work, we build a generative model that learns to capture such object intrinsics from a single image, such as a photo of a bouquet. Such an image includes multiple instances of an object type. These instances all share the same intrinsics, but appear different due to a combination of variance within these intrinsics and differences in extrinsic factors, such as pose and illumination. Experiments show that our model successfully learns object intrinsics (distribution of geometry, texture, and material) for a wide range of objects, 
each from a single Internet image. Our method achieves superior results on multiple downstream tasks, including intrinsic image decomposition, shape and image generation, view synthesis, and relighting.
\vspace{-5pt}
\end{abstract}

\begin{formal}
\emph{``Nature!... Each of her works has an essence of its own; each of her phenomena a special characterisation; and yet their diversity is in unity.''} -- Georg Christoph Tobler
\end{formal}

\section{Introduction}

\footnotetext{``People where you live ... grow five thousand roses in one garden ... And yet what they're looking for could be found in a single rose.'' --- Quote from \emph{The Little Prince} by Antoine de Saint-Exupéry.}

The bouquet in \Cref{fig:teaser} contains many roses. Although each rose has different pixel values, we recognize them as individual instances of the same type of object. 
Such understanding is based on the fact that these instances share the same \emph{object intrinsics}---the distributions of 
geometry, texture, and material that characterize a rose.
The difference in appearance arises from both the variance within these distributions and extrinsic factors such as object pose and environment lighting. Understanding these aspects allows us to
imagine and draw additional, new instances of roses with varying shapes, poses, and illumination.

In this work, our goal is to build a model that captures such object intrinsics from \emph{a single image}, and to use this model for shape and image generation under novel viewpoints and illumination conditions, as illustrated in \Cref{fig:teaser}. 

This problem is challenging for three reasons. First, we only have a single image. This makes our work fundamentally different from existing works on 3D-aware image generation models~\cite{nguyen2019hologan, niemeyer2021giraffe, piGAN2021, chan2022eg}, which typically require a large dataset of thousands of instances for training. 
In comparison, the single image contains at most a few dozen instances, making the inference problem highly under-constrained.

Second, these already limited instances may vary significantly in pixel values. This is because they have different poses and illumination conditions, but neither of these factors is annotated or known.
We also cannot resort to existing tools for pose estimation based on structure from motion,
such as COLMAP~\cite{schoenberger2016sfm}, because the appearance variations violate the assumptions of epipolar geometry.

Finally, the object intrinsics we aim to infer are probabilistic, not deterministic: no two roses in the natural world are identical, and we want to capture a distribution of their 
geometry,  texture, and material
to exploit the underlying multi-view information. This is therefore in stark contrast to existing multi-view reconstruction or neural rendering methods for a fixed object or scene~\cite{mildenhall2020nerf, wang2021neus, meng2021gnerf}. 

These challenges all come down to having a large hypothesis space for this highly under-constrained problem, with very limited visual observations. Our solution to address these challenges is to design a model with its inductive biases guided by \emph{object intrinsics}. Such guidance is two-fold: first, the instances we aim to present share the same object intrinsics, or the same distribution of geometry, texture, and material; second, these intrinsic properties are not isolated, but interweaved in a specific way as 
defined by
a rendering engine and, fundamentally, by the physical world. 

Specifically, our model takes the single input image and learns a neural representation of the distribution over 3D shape, surface albedo, and shininess of the object, factoring out pose and lighting variations, based on a set of instance masks and
a given pose distribution of the instances. This explicit, physically-grounded disentanglement helps us explain the instances in a compact manner, and enables the model to learn object intrinsics without overfitting the limited observations from only a single image. 

The resulting model enables a range of applications. 
For example, random sampling from the learned \objint generates novel instances with different identities from the input.
By modifying extrinsic factors, the synthesized instances can be rendered from novel viewpoints or relit with different lighting configurations.

Our contributions are three-fold:
\begin{enumerate}
    \item We propose the problem of recovering object intrinsics, including both 3D geometry, texture, and material properties, from just a single image of a few instances with instance masks.
    \item We design a generative framework that effectively learns such object intrinsics.
    \item Through extensive evaluations, we show that the model achieves superior results in shape reconstruction and generation, novel view synthesis, and relighting. 
\end{enumerate}

\begin{table}[t]
\centering
\small
\begin{tabular}{lcccc}
\toprule

&\makecell{Object\\ Variance}
&\makecell{Unknown \\Poses}
&\makecell{Re-\\lighting}
&\makecell{3D-\\Aware}
\\ \midrule

SinGAN~\cite{shaham2019singan} & $\checkmark$ &$\checkmark$  &&\\
NeRF~\cite{mildenhall2020nerf} &  &  && $\checkmark$  \\
D-NeRF~\cite{pumarola2020d} & $\checkmark$  &  &  &   $\checkmark$  \\
GNeRF~\cite{meng2021gnerf} && $\checkmark$ &&  $\checkmark$ \\
Neural-PIL~\cite{boss2021neuralpil}  &  &  & $\checkmark$    & $\checkmark$  \\
NeRD~\cite{boss2021nerd} &  &  & $\checkmark$    & $\checkmark$  \\
EG3D~\cite{chan2022eg} & $\checkmark$ &  && $\checkmark$ \\
Ours & $\checkmark$ & $\checkmark$ & $\checkmark$ & $\checkmark$\\

\bottomrule
\end{tabular}

\caption{Comparisons with prior works. 
Unlike existing 3D-aware generative models, our method learns from a very limited number of observations. 
Unlike multi-view reconstruction methods, our method models variance among observations from training inputs. 
}
\vspace{-1.2em}
\label{tbl:related}
\end{table}

\section{Related Work}

\paragraph{Generative Models with Limited Data.}
Training generative models~\cite{karras2019style,orel2021stylesdf,chan2022eg} typically requires large datasets of thousands of images. ADA~\cite{Karras2020ada} proposes a differentiable data augmentation technique that targets a more data-limited regime, but still in the magnitude of a thousand. Several internal learning methods propose to exploit statistics of local image regions and are able to learn a generative model from a single image for image synthesis~\cite{shaham2019singan,shocher2019ingan,granot2022drop,kulikov2022sinddm,wang2022sindiffusion}, or learn from a single video for video synthesis~\cite{haim2022diverse}, but these methods typically do not explicitly reconstruct 3D geometry. Recently, SinGRAV~\cite{wang2022singrav} applies the multi-scale learning strategy from the internal learning literature to tackle the task of 3D-consistent scene generation, but training the model requires hundreds of observations for each scene. 3inGAN~\cite{karnewar20223ingan} proposes a generative model for scenes with self-similar patterns and requires multiple observations for the same scene with ground-truth camera poses. 

\paragraph{Intrinsics Image Decomposition.}
To disentangle \objint from extrinsic factors, we seek to find the distribution of true surface color for a type of object. This is closely connected to the classic task of intrinsics image decomposition, where an input image is decomposed into an albedo and shading map. This is a highly under-constraint problem, and 
prior works tackle the task of intrinsics image decomposition from a single image with heuristics assumptions such as global sparsity of the reflectance~\cite{barron2012color,BarronTPAMI2015, laffont2012coherent}. Several learning-based approaches~\cite{liu2020unsupervised,YuSelfRelight20,wu2021derender,wimbauer2022derender} adapt the heuristics as regularizations during training.
Unlike these methods, we seek for training regularizations by exploiting the underlying multiview signals among observations. 
\paragraph{Neural Volumetric Rendering.}
Several recent methods~\cite{mildenhall2020nerf,wang2021neus,yariv2020multiview,niemeyer2020differentiable} use neural volumetric rendering to learn implicit scene representations for 3D reconstruction tasks. 
In this work, we integrate an albedo field with the recent NeuS~\cite{wang2021neus} representation to capture \objint.
These can be further extended to recover not only geometry, but also material properties and illuminations from scenes~\cite{zhang2021nerfactor,boss2021nerd,boss2021neuralpil,zhang2021physg,verbin2022ref}.
However, they typically require densely captured multi-view observations for a single scene, and do not generalize across different instances as ours. 
Several methods~\cite{du2021nerflow,pumarola2020d} extend the NeRF representation to handle variance among observations, but all these methods require ground-truth camera poses, while our method does not. 

\Cref{tbl:related} gives an overall comparison of our method with prior works. 

\begin{figure}[t]

  \centering
  \includegraphics[width=\linewidth]{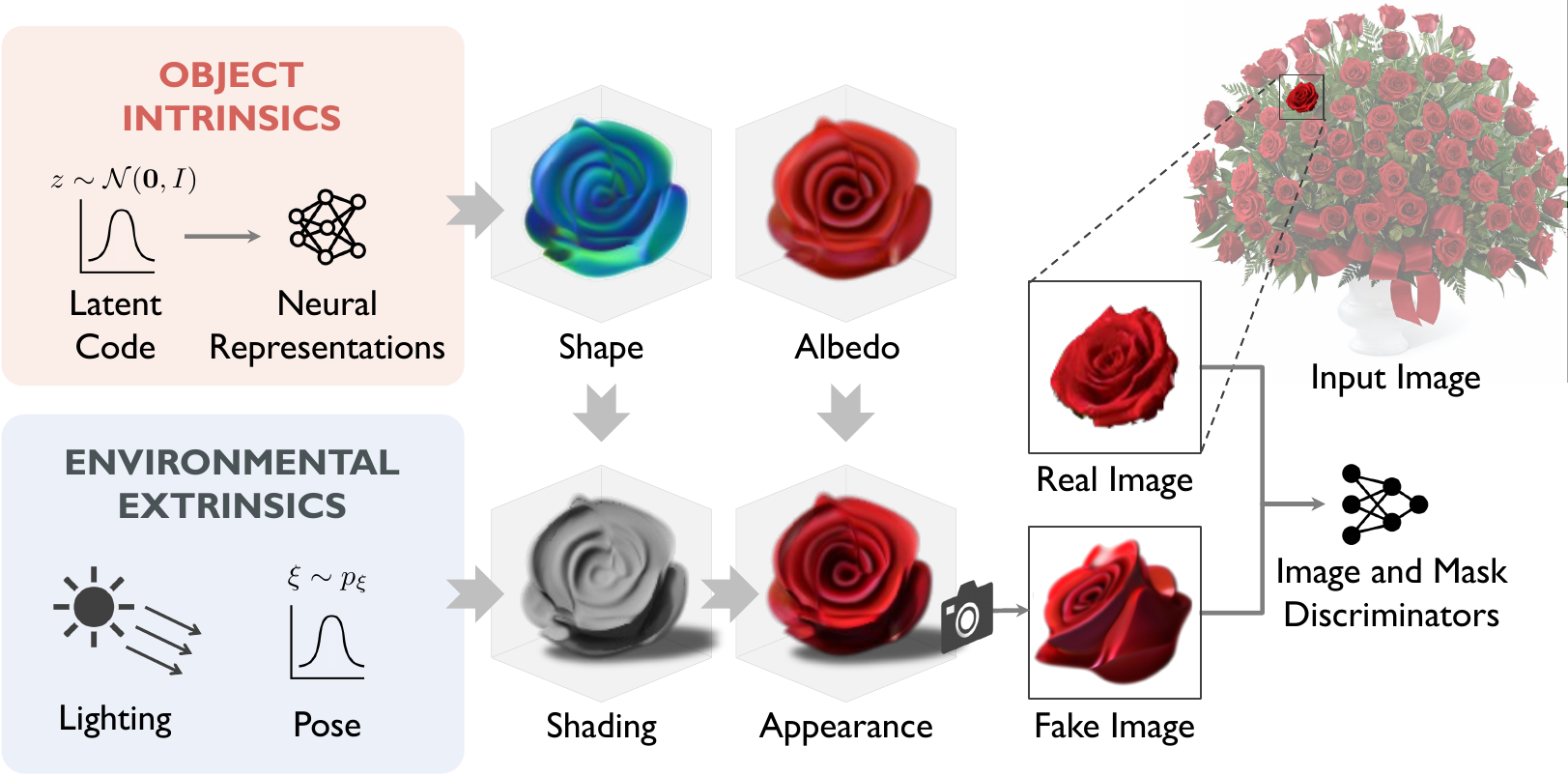}
  \caption{Model overview. 
  We propose a generative model that recovers the \objint, including 3D shape and albedo, from a single input image with multiple similar object instances with instance masks.
  To synthesize an image, we sample from the learned \objint (orange box) to obtain the shape and albedo for a specific instance, whose identity is controlled by an underlying latent space. 
  Then, environmental extrinsics (blue box) are incorporated in the forward rendering procedure to obtain shading and appearance for the instance. 
  Finally, the 3D representation for appearance is used to render images in 2D under arbitrary viewpoints.
  These synthesized images are then used, along with the real examples from the input image, in a generative adversarial framework to learn the \objint.
  }
  \vspace{-1em}
  \label{fig:model}
\end{figure}
\section{Method}

Given a single 2D image $I$ containing a few instances of the same type of object, together with 
$K$ instance masks $\{M_k\}_{k=1}^K$ and a roughly estimated pose distribution $p_{\xi}$ over the instances, our goal is to learn a generative model that captures the \objint of that object category, namely the distributions of 3D shape and surface albedo.
We do not rely on any other geometric annotations, such as the exact pose of each individual instance, or a template shape.

As illustrated in \Cref{fig:model},
for training, we sample a 3D neural representation from the \objint (\cref{subsect:representations}), and render 2D images with a physics-based rendering engine (\cref{subsect:rendering}), taking into account the environmental extrinsics. 
The \objint is learned through an adversarial training framework (\cref{subsect:adversarial}), matching the distribution of rendered 2D images with the distribution of masked instances from the input image.

\subsection{Representations}
\label{subsect:representations}

We model each factor of the object intrinsics using  a neural representation, including geometry, texture, and material.
In order to model the variations among instances, the networks are conditioned on a latent vector $\mathbf{z} \in \mathbb{R}^d$ ($d=64$) drawn from a standard multivariate normal distribution.

To represent the geometry, we adopt a recent neural field representation based on NeuS~\cite{wang2021neus}, which parameterizes the 3D shape using a Signed Distance Function (SDF).
Specifically, $f_\theta: \mathbb{R}^3 \times \mathbb{R}^d \rightarrow \mathbb{R}$ maps a spatial location $\mathbf{x} \in \mathbb{R}^3$ and a latent vector $\mathbf{z} \in  \mathbb{R}^d$ to the signed distance of $\mathbf{x}$ from the object surface, where $\theta$ denotes network parameters.
To simplify notations, $\mathbf{z}$ is omitted from now on. 
With this SDF $f_\theta$, an object surface can be expressed as the zero level set:
\begin{equation}
   \label{eqn:surface}
    \mathcal{S}_\theta = \{\mathbf{x}\in\mathbb{R}^3\mid f_\theta(\mathbf{x}) = 0\}.
\end{equation}
To encourage the output of $f_\theta$ to be a signed distance function, we impose an Eikonal regularizer~\cite{gropp2020implicit}:
\begin{equation}
    \label{eqn:loss_reg}
    \mathcal{L}_\text{eikonal}(\theta) = \sum_{\mathbf{x} \in \mathbb{R}^3}(\|\nabla f_\theta(\mathbf{x})\|_2 - 1)^2.
\end{equation}
The surface normal $\mathbf{n}_\theta: \mathbb{R}^3\times \mathbb{R}^d\rightarrow\mathbb{R}^3$ hence can be derived from the gradient of the SDF $\nabla_{\mathbf{x}} f_\theta$ through automatic differentiation.

To represent the texture, we use an albedo network $a_\psi: \mathbb{R}^3 \times \mathbb{R}^d \rightarrow \mathbb{R}^3$ that predicts the RGB value of albedo associated with a spatial point $\mathbf{x} \in \mathbb{R}^3$ and a latent code $\mathbf{z}\in \mathbb{R}^d$, where $\psi$ denotes the trainable parameters.

To model the surface material, we optimize a shininess scalar $\alpha\in\mathbb{R}$ using a Phong illumination model, described next in \Cref{eqn:intrinsics}.

\subsection{Forward Rendering}
\label{subsect:rendering}
\paragraph{Lighting and Shading.}
We use a Phong illumination model under the effect of a dominant directional light source.

Let $\mathbf{l}_\text{global}\in\mathbb{S}^2$ be the light direction, $k_d, k_a, k_s\in\mathbb{R}$ the diffuse, ambient, and specular coefficients, and $\alpha\in\mathbb{R}$ the shininess value.
An instance with pose $\xi\in \mathrm{SE}(3)$ receives an incoming light with direction $\mathbf{l} = \xi \mathbf{l}_\text{global}$ in its canonical frame.
The radiance color at spatial location $\mathbf{x}\in\mathbb{R}^3$ with viewing direction $\mathbf{v}\in \mathbb{R}^3$ is computed as
\begin{equation}
\label{eqn:intrinsics}
    c(\mathbf{x}) = s_\theta(\mathbf{x}) a_\psi(\mathbf{x}) + k_s\max\{ (\mathbf{r}_\theta(\mathbf{x}, \mathbf{l}) \cdot\mathbf{v})^{\alpha}, 0\},
\end{equation}
where the diffuse component is given by
\begin{equation}
    \label{eqn:shading}
    s_\theta(\mathbf{x}) = k_d \max\{\mathbf{n}_\theta(\mathbf{x}) \cdot \mathbf{l}, 0\}+ k_a,
\end{equation}
and $\mathbf{r}_\theta$ is the reflection of $\mathbf{l}$ with normal direction $\mathbf{n}_\theta(\mathbf{x})$. 

For initialization, $k_a = 1/3, k_d = 2/3, k_s = 0, \alpha = 10$, and $\mathbf{l}_\text{global}$ is estimated for each input image. These parameters are jointly optimized during training. 
$k_a, k_d$ are reparameterized as $k_a = S(\beta), k_d = 1 - S(\beta)$, where $\beta\in\mathbb{R}$ and $S(\beta)=1/(1+e^{-\beta})$ is the sigmoid function.

\paragraph{Neural Volume Rendering.}
\label{sec:volume}
Next, we describe in detail the rendering operation, denoted as $\mathcal{R}$.
Without loss of generality, the camera pose is fixed to be the identity. We assume access to an approximate prior pose distribution $p_\xi$, from which instance poses $\xi \in \mathrm{SE}(3)$ are sampled during training.
For each pixel to be rendered, we cast a ray from the camera center, which is set to be the origin, through the pixel. 
Points on the ray $\mathbf{r}(r) = r\mathbf{v}$ with viewing direction $\mathbf{v}$ are transformed to the canonical object frame $\xi \mathbf{r}(r)$ before querying the shape and albedo networks.

The final color of the pixel $C(\mathbf{r}; \xi)$ is defined as:
\begin{equation}
    \label{eqn:neus}
    C(\mathbf{r}; \xi)= \int_0^{+\infty} w(r) c(\xi \mathbf{r}(r)) \;dr,
\end{equation}
\begin{equation}
    \label{eqn:neus_density}
    \text{where} \quad w(r) = \frac{\phi_s(f_\theta(\xi \mathbf{r}(r)))}{\int_0^{+\infty} \phi_s(f_\theta(\xi \mathbf{r}(u)))\;du}.
\end{equation}
The weight function $w(r)$ is same as derived in NeuS~\cite{wang2021neus} to ensure unbiased surface reconstruction, where $\phi_s(x) = se^{-sx}/(1+e^{-sx})^2$ is the logistic density distribution with a global scaling parameter $s$.

The above integration is estimated as the sum of radiance colors of discrete samples along the ray weighted by densities. Specifically, for each ray, $N=16$ points are sampled with training-time noise, and then $N_\text{importance}=4$ points are obtained via importance sampling following the coarse-to-fine strategy in NeRF~\cite{mildenhall2020nerf}.

\subsection{Generative Adversarial Training}
\label{subsect:adversarial}
As we do not assume known object instance poses,
the neural field networks cannot be directly optimized using a reconstruction loss, as typically done in NeRFs~\cite{mildenhall2020nerf}.
Estimating the instance pose is a challenging problem since variations among instances make it difficult to establish correspondences.
Instead, we use an adversarial network~\cite{goodfellow2014generative} to train the implicit representation from \cref{subsect:rendering}. Specifically, we train an image discriminator $D_\eta$ which receives image crops around instances from real or fake scenes. 

\paragraph{Image Crops.}
\label{sec:crop}
Rendering all object instances in the full scene with the volume rendering operation $\mathcal{R}$ from \cref{sec:volume} is memory-inefficient, since each instance typically only occupies a small region in the image plane. Instead, the generator renders only a crop around the object. To decide the crop offset, we project a unit sphere co-centered with the object to the image plane, and only render pixels contained in the axis-aligned bounding box of the projection. 
The crop-based rendering strategy prevents computing whether each ray from the camera intersects with the rendered object, which requires extra queries of the shape network. 

Correspondingly, the discriminator receives image crops as inputs. Real image crops are obtained from $\{I_k\}_{k=1}^K$ for $K$ instances in the input image $I$, where each $I_k$ is the center crop of $I \odot M_k$. The crop size is set to be the maximum bounding box size among all instances. 

\paragraph{Scale and Translation Augmentations.}
We design our framework such that the generator is 3D-consistent and the discriminator is 2D-scale- and 2D-translation-invariant. 
Specifically, we make the discriminator robust to the distribution shift of scale and translation between real and fake data distribution by applying random translation and random scaling to the 2D image crops as data augmentation. 

The augmentation is also used in Adaptive Discriminator Augmentation (ADA)~\cite{Karras2020ada}. However, since the generator used in ADA does not have a 3D representation, it requires an augmentation probability $p < 1$ together with a tuning schedule to prevent the distorted distribution of augmented data from being leaked to the generator. The same technique is directly adopted in previous work with 3D-aware generators such as EG3D~\cite{chan2022eg}, also with $p < 1$. In contrast, we exploit the fact that a 3D-aware physical rendering procedure enforces geometric consistency by design and, therefore, would not suffer from distribution leakage with 2D augmentations. Because of this, we use $p = 1$ in all experiments. 

The augmentation stabilizes training given the limited amount of data and improves robustness to the approximation error between the estimated and the real, unknown pose distribution for instances observed in the input image.

\paragraph{Discriminator Design.}
To stabilize training, the discriminator predicts the pose of the instance used to generate fake image crops. This regularization term is defined as:  
\begin{equation}
    \mathcal{L}_\text{pose} = \| g_\text{GS} (\hat{\xi}_\text{rot}) - g_\text{GS}(\xi_\text{rot})\|_2^2, 
\end{equation}
where $\xi_\text{rot}$ is the rotations sampled at generation, and $\hat{\xi}_\text{rot}$ the one predicted from $D_\eta$, 
and $g_\text{GS}$ is the Gram-Schmidt process that maps a $\mathrm{SO}(3)$ rotations to a $6D$ embedding by dropping the last column, following \cite{zhou2019continuity}. 

In addition to $D_\eta$, we use a second discriminator $D_{\eta_\text{mask}}$ for masks, which receives cropped masks from the generator and input instance masks. Empirically, we found that the additional discriminator improves training stability. 

\paragraph{Training Objective.}
Similar to GIRAFFE~\cite{niemeyer2021giraffe}, we use the binary cross entropy loss as the adversarial training objective, with a regularization term on gradients of the discriminator:
\begin{equation}
\begin{aligned}
    &\mathcal{L}_\text{adv}(\theta, \psi, \eta) 
   = 
    \mathbb{E}_{\xi\sim p_\xi}\left[f(D_\eta(\mathcal{R}(\xi))\right] \\
    +& 
    \mathbb{E}_{k\sim \{1, \cdots, K\}}\left[f(-D_\eta(I_k)) - \lambda_\text{reg} \|\nabla D_\eta(I_k)\|^2\right],
\end{aligned}
    \label{eqn:loss_adv}
\end{equation}
where $\mathcal{R}(\xi)$ is the forward rendering operation in \cref{subsect:rendering}.

The final training objective, thus, comprises four terms:
\begin{equation}
\begin{aligned}
    \label{eqn:loss}
        &\mathcal{L}(\theta, \psi, \eta) 
        =
        \mathcal{L}_\text{adv}(\theta, \psi, \eta)
        + \lambda_\text{mask}\mathcal{L}_\text{adv}(\theta, \psi, \eta_\text{mask})\\
        & + \lambda_\text{pose}\mathcal{L}_\text{pose}(\eta)
        +\lambda_\text{eikonal}\mathcal{L}_\text{eikonal}(\theta).
\end{aligned}
\end{equation}
\subsection{Training Details}
Across all experiments, we use a resolution of $128 \times 128$ at training, and augment the background uniformly with a random color on both fake and real images. 
Weights of the loss terms in \cref{eqn:loss} 
are specified as $\lambda_\text{reg} = 10, \lambda_\text{eikonal} = 10, \lambda_\text{pose} = 1$, and $\lambda_\text{mask} = 0.1$ or $1$ depending on the training image. 
We adopt the generator backbone from StyleSDF~\cite{orel2021stylesdf} using SIREN~\cite{sitzmann2019siren} as the activation function. The backbone for both discriminators is adapted from GIRAFFE~\cite{niemeyer2021giraffe}.
The detailed architecture is specified in the supplement.
The shape network is initialized as a unit sphere centered at the origin. 
We use an Adam~\cite{kingma2014adam} optimizer with learning rates $2e-5$ and $1e-4$ for the generator and discriminators, respectively. 

\begin{figure*}[t]
  \centering
\includegraphics[width=\linewidth]{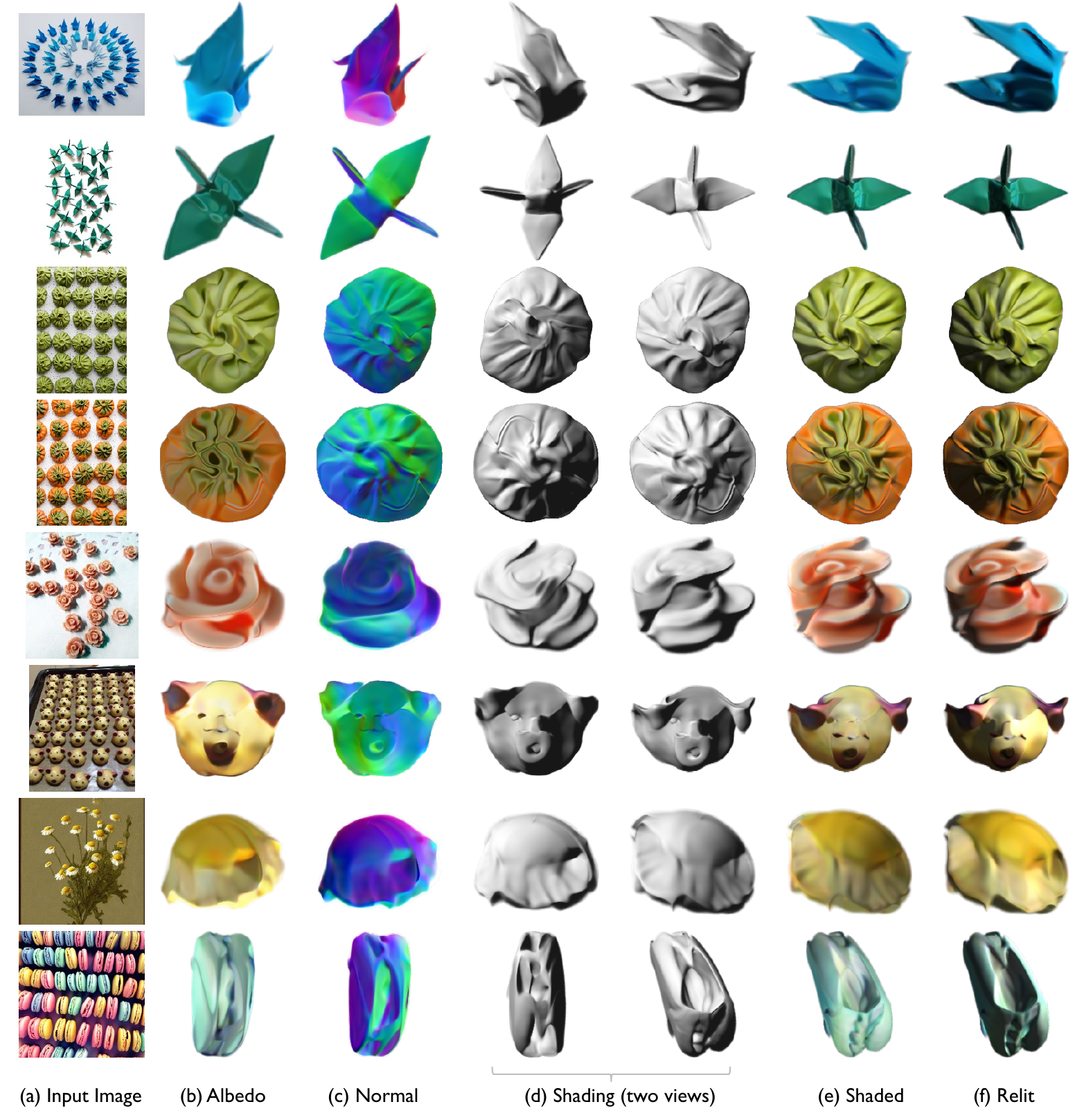}
\caption{Learning from images in the wild. 
Given a single 2D image containing dozens of similar object instances with masks, our model can recover a distribution of 3D shape and albedo from observations of the instances. 
We sample from the learnt distribution to obtain albedo and normal for a specific instance, as shown in column (b-c). Two columns in (d) show two different views for the same instance. 
At test time, our model can synthesize instances under novel views shown in (e) and novel lighting conditions shown in (f).
}
\vspace{-1em}
\label{fig:wild}
\end{figure*}

\begin{figure}[t]
  \centering
\includegraphics[width=\linewidth]{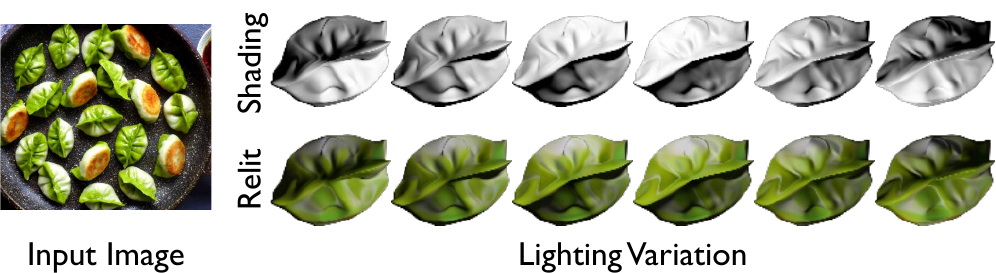}
  \caption{Results for test-time relighting. The 6 columns show renderings with different lighting conditions unseen during training.}
\label{fig:light}
\end{figure}
\begin{figure}[t]
  \centering
\includegraphics[width=\linewidth]{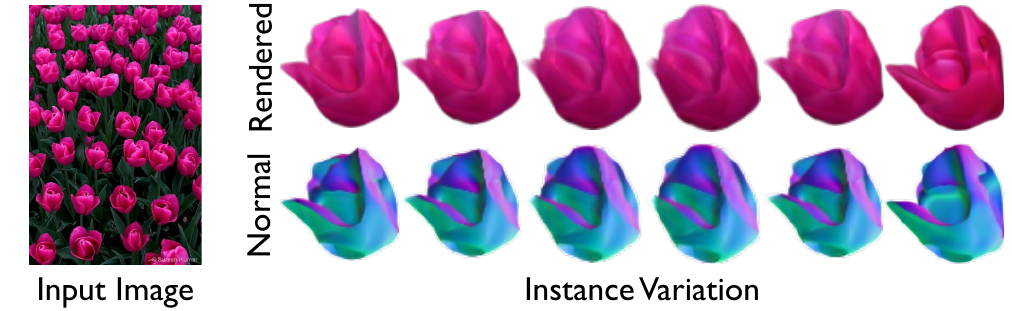}
  \caption{Results of interpolation in the latent space. From left to right, each column of images corresponds to an instance with a specific latent code interpolated between two latent vectors. Instances from all columns are rendered with the same pose. }
  \vspace{-1em}
\label{fig:latent}
\end{figure}

\section{Experiments}

We test our method on a diverse set of real-world images and evaluate extensively on both real scanned objects and synthetic objects.
Experiments show that our proposed method successfully recovers object intrinsics from in-the-wild images, and achieves superior 3D reconstruction and synthesis results compared to prior art.

\subsection{Learning from in-the-Wild Images}
\label{subsect:wild}
It is common to see online images with a group of similar objects placed in a scene. These images have unknown camera intrinsics or extrinsics, unknown object poses, and uncontrolled lighting conditions, all posing significant challenges to the task of 3D reconstruction and generative modeling. We test whether our method can robustly recover the geometry and texture of objects and capture the variation among object instances under this challenging setting. 

\paragraph{Dataset.} 
We collect a set of 11 in-the-wild images containing multiple similar object instances, shown in \Cref{fig:teaser,fig:wild,fig:latent,fig:light}. Out of the 11 images, nine are Internet photos, and the other two are generated by Stable Diffusion~\cite{rombach2021highresolution} (the last two rows of \Cref{fig:wild}). These images altogether cover a diverse range of object categories. 

To obtain foreground masks, we use a pre-trained model from $U^2$-Net~\cite{qin2020u2}, or manually segment the immages when $U^2$-Net fails. We then run a connected-components-finding algorithm~\cite{bolelli2019spaghetti} on foreground pixels to obtain a set of instance masks. 
Examples of cropped real instances are shown in supp. mat. 
For each image, the camera is estimated as a pinhole camera with a field-of-view of $10\degree$ after cropping with the obtained instance masks. The pose distribution is estimated as randomly distributed either on a sphere, or across a 3D plane while remaining visible in the image, optionally with random rotations around an axis in the local object frame depending on the input image. 

\paragraph{Results.}
We show qualitative results in \Cref{fig:wild}. For each training image, given only a handful of observations of masked instances, our method faithfully recovers the 3D geometry and texture of the object category. 

Furthermore, the physical disentanglement of environmental extrinsics and \objint allows the proposed method to perform several inference tasks.
Specifically, changes to poses or lighting, both components of the extrinsic configuration, correspond to novel-view synthesis (\Cref{fig:wild}) and test-time relighting (\Cref{fig:light}), respectively. 
In addition, different samples from the learned \objint (\Cref{fig:latent}) correspond to different instance identities, making it possible to synthesize novel instances unseen in the original image. In \Cref{fig:latent}, the shape and texture of generated instances vary across latent vectors, but poses remain relatively stable, showing that the model disentangles poses from the identity.
See supp. mat. for more visualizations.

\begin{table}[t]
\centering
\footnotesize
\begin{tabular}{lccccc}
\toprule
 & \multicolumn{1}{c}{Depth}
& \multicolumn{3}{c}{Appearance} 
 & \multicolumn{1}{c}{Generation}
 \\ 
 \cmidrule(r){2-2}\cmidrule(r){3-5}
\cmidrule(r){6-6}
 & \multicolumn{1}{c}{MSE$\downarrow$} 
 & \multicolumn{1}{c}{SSIM$\uparrow$} & \multicolumn{1}{c}{PSNR$\uparrow$} &\multicolumn{1}{c}{LPIPS$\downarrow$} 
 & \multicolumn{1}{c}{FID$\downarrow$} 
 \\ \midrule
GNeRF~\cite{meng2021gnerf}&0.12 &5.39 & 0.29&0.62&353.61\\
Ours &\textbf{0.01} & \textbf{19.89} & \textbf{0.75} & \textbf{0.13} &\textbf{204.63} \\\bottomrule
\end{tabular}
\caption{Results on shape and appearance reconstruction averaged over three real-world-captured scenes, evaluated on held-out instances. Compared to GNeRF, our method achieves significantly lower reconstruction error for both geometry and appearance, and better image generation quality measured by FID.}
\label{tbl:real}
\vspace{-1.2em}
\end{table}

\begin{table*}[t]
\centering
\small
\begin{tabular}{lccccccccccc}
\toprule

 & \multicolumn{1}{c}{Normal} 
 & \multicolumn{1}{c}{Depth} 
 & \multicolumn{3}{c}{Albedo} & \multicolumn{3}{c}{View Synthesis} & \multicolumn{3}{c}{Relighting}\\ 
\cmidrule(r){2-2}
\cmidrule(r){3-3}
\cmidrule(r){4-6}
\cmidrule(r){7-9}
\cmidrule(r){10-12}
 & Angle($\degree$)$\downarrow$ & MSE$\downarrow$ 
 & PSNR$\uparrow$ & SSIM$\uparrow$ & LPIPS$\downarrow$ 
 & PSNR$\uparrow$ & SSIM$\uparrow$ & LPIPS$\downarrow$ 
 & PSNR$\uparrow$ & SSIM$\uparrow$ & LPIPS$\downarrow$ 
 \\ \midrule

NeRF$^*$~\cite{mildenhall2020nerf} & \na & \textbf{0.98} & \na & \na & \na & \textbf{29.95} & \textbf{0.94} & \textbf{0.06} & \na & \na & \na \\
NeRD$^*$~\cite{boss2021nerd} & 79.64 & 30.85 & 13.63 & 0.83 & 0.21 & 18.47 & 0.84 & 0.18 & 18.22 & 0.84 & 0.19 \\
Neural-PIL$^*$~\cite{boss2021neuralpil} & 69.50 & 31.03 & 18.24 & 0.85 & 0.15 & 25.42 & 0.88 & 0.10 & 24.86 & 0.87 & 0.11 \\ \midrule
GNeRF~\cite{meng2021gnerf} & \na & 4.07 & \na & \na & \na & 22.85 & 0.83 & 0.19 & \na & \na & \na \\
Ours & \textbf{22.69} & \textbf{1.10} & \textbf{22.42} & \textbf{0.87} & \textbf{0.10} & \textbf{29.13} & \textbf{0.93} & \textbf{0.04} & \textbf{25.94} & \textbf{0.91} & \textbf{0.06}\\

\bottomrule
\end{tabular}
\caption{Results on synthetic data. Our method yields better or comparable reconstruction quality compared to all baseline methods, including those with access to ground truth poses (denoted as $^*$). It also achieves superior results on albedo decomposition and test-time relighting across all metrics.
}
\vspace{-1.2em}
\label{tbl:synthetic}
\end{table*}
\subsection{Shape Evaluation on Real Captured Scenes}
\label{subsect:real}

\begin{figure}[t]
  \centering

  \includegraphics[width=\linewidth]{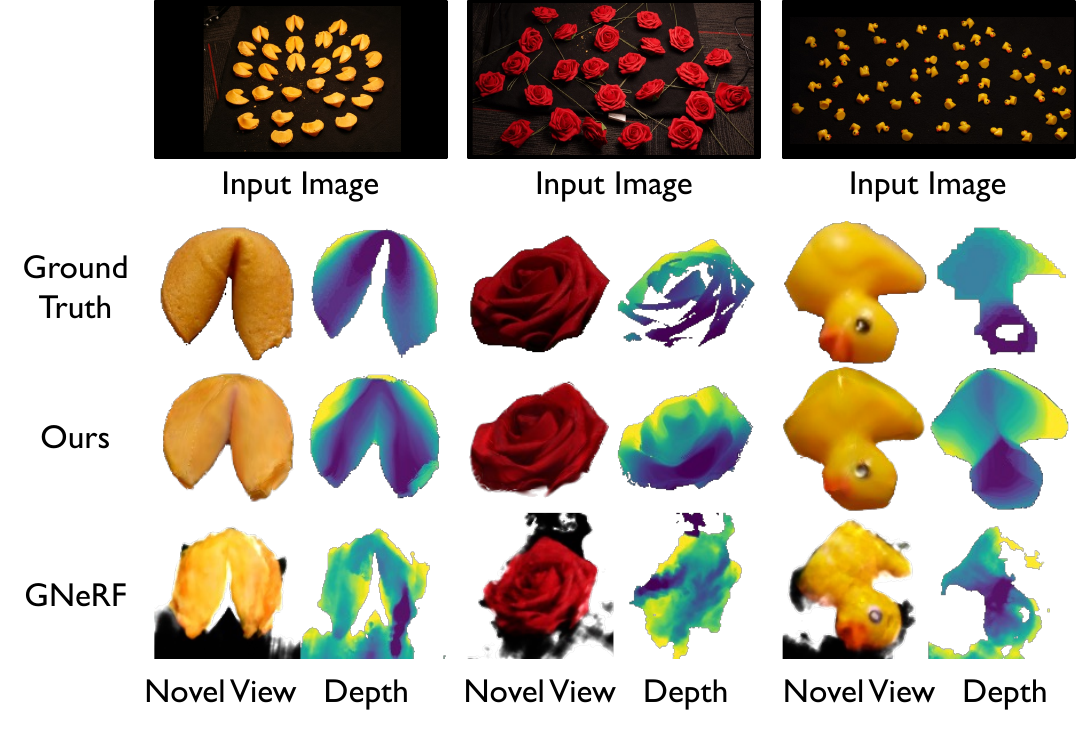}
  \vspace{-1.2em}
  \caption{Qualitative results on real-world-captured scenes. Our method can reconstruct the geometry and appearance of novel instances held out from training more faithfully compared to the baseline method. }
  \label{fig:real}
  \vspace{-1.2em}

\end{figure}

\paragraph{Dataset.}
In order to quantitatively evaluate the reconstruction quality of the proposed method, we collect three scenes shown in \Cref{fig:real}, where each scene contains 25-64 object instances of the same category. 
For each scene, we additionally capture an image held out from training, which contains three object instances in a different layout. We 3D-scan the test scene to obtain the ground-truth depth maps. The preprocessing procedure is the same as described in \Cref{subsect:wild}.
We approximate the prior pose distribution as follows: instances are randomly placed on the ground with a random rotation around the up axis in their local frames, and the camera has an elevation of $45\degree$ relative to the ground. 

\paragraph{Metrics.}
We measure the quality of shape reconstruction using the scale-invariant mean squared error (MSE) on depth map predictions, defined as 
$\mathcal{L}(\mathbf{x}, \hat{\mathbf{x}}) := \min_{\hat{\alpha} \in \mathbb{R}} \|\mathbf{x}-\hat{\alpha}\hat{\mathbf{x}}\|^2$ following~\cite{eigen2014depth}.
The image reconstruction quality is measured by the Peak Signal-to-Noise Ratio (PSNR), the Structural Similarity Index Measure (SSIM)~\cite{wang2004image}, and the Learned Perceptual Image Patch Similarity (LPIPS)~\cite{zhang2018unreasonable}.

To evaluate the quality of image generation, we report Frechet
Inception Distance (FID)~\cite{heusel2017gans}, which measures the statistical difference of distributions of real and fake samples projected to the feature space of a neural network pre-trained on ImageNet~\cite{deng09imagenet}. 
In our case, the real and fake distributions are formed by all real image crops from the training scene, and synthesized image crops rendered with random poses from the prior pose distribution, respectively. 

\paragraph{Baselines.}
Most prior methods for 3D reconstruction are either designed to learn from multiple views of the \emph{same} object instance or scene with no variations~\cite{mildenhall2020nerf}, or require a large dataset of on the order of 1k--10k images~\cite{orel2021stylesdf,chan2022eg}. 
Given the limited number of observations available and with unknown poses, the method closest to our setting is GNeRF~\cite{meng2021gnerf}. Given a prior pose distribution and a collection of multi-view images of a scene, GNeRF jointly estimates per-image camera poses and optimizes for a 3D representation by iteratively training a pose network and a NeRF~\cite{park2021nerfies}-based network with a discriminator loss. 
To train GNeRF, we use the image crops as multiple views of the same object.

Other multi-view reconstruction methods, such asNeRF~\cite{mildenhall2020nerf} and NeuS~\cite{wang2021neus}, require camera poses for each training image
which are typically estimated with COLMAP~\cite{schoenberger2016sfm}. Variations among instances and different lighting configurations make pixel-based matching very challenging on the converted multi-view data, and we found that COLMAP does not converge for 
any of the three scenes, making these baseline methods inapplicable.

\paragraph{Implementation Details.}
We train our method for 100k iterations for all scenes, and use the same pose distribution, as specified above, for both our method and the baseline. 

During inference, for each held-out instance, we use GAN-inversion to find the pose and latent code associated with the instance. We freeze the model after training, and first randomly sample $1{,}000$ poses from the prior pose distribution with a fixed latent code averaged over $10{,}000$ samples in the latent space. The top-$5$ poses ranked by LPIPS errors are selected, and the latent code gets updated for $2{,}000$ gradient steps for each pose, with an Adam~\cite{kingma2014adam} optimizer of learning rate $4e-3$. The pose with the lowest LPIPS error after gradient updates is used for evaluation. 
For GNeRF~\cite{meng2021gnerf}, which is not conditioned on a latent code, we randomly sample $5{,}000$ poses from the prior distribution, and use the one with the lowest LPIPS error for evaluation.

\paragraph{Results.}
Both quantitatively (\Cref{tbl:real}) and qualitatively (\Cref{fig:real}), our method achieves higher-fidelity synthesis results compared to the baseline method across all three scenes and all metrics. Our neural representations for geometry and appearance capture the distribution across observed instances as opposed to optimizing for one single instance as in GNeRF, which allows our method to better reconstruct novel instances unseen during training. 

\begin{figure}[t]
  \centering
  \includegraphics[width=\linewidth]{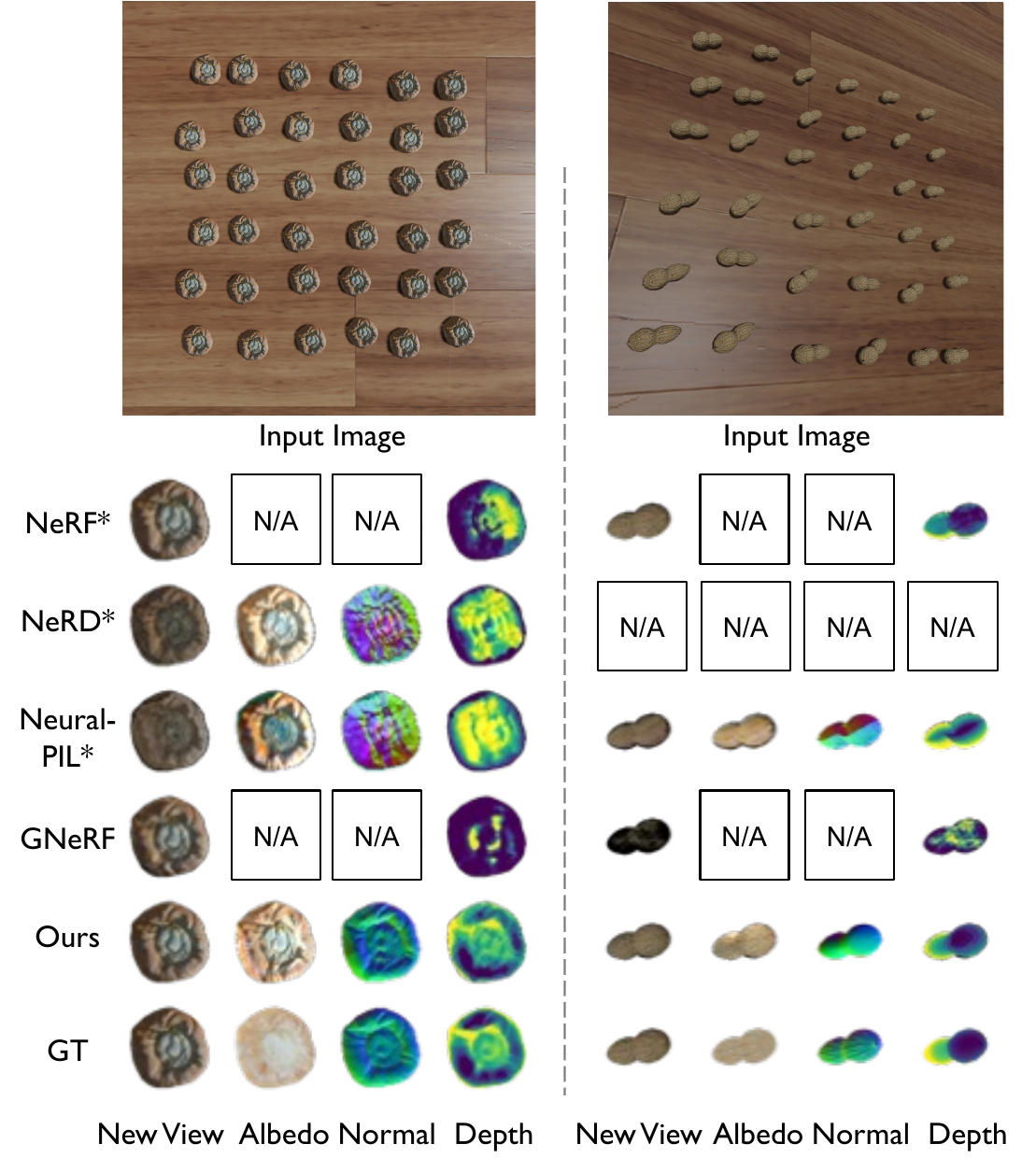}
  \caption{Results of intrinsic decomposition on synthetic datasets. NeRD~\cite{boss2021nerd} does not converge on the second dataset, denoted as N/A. 
  GNeRF~\cite{meng2021gnerf} does not decompose image intrinsics and does not directly predict normal, also denoted as N/A. $*$ denotes methods provided with ground-truth poses.
  }
  \vspace{-1em}
\label{fig:synthetic}
\end{figure}
\subsection{Intrinsics Evaluation on Synthetic Scenes}

\paragraph{Dataset.}
To test whether our model can recover object intrinsic properties such as albedo, we use synthetic data for evaluation, as we do not have ground-truth albedo for real objects. 

Our dataset has four scenes containing 36 instances each, rendered from two assets~\cite{blenderbun,blenderpeanut} with two pose and lighting configurations. All instances are located on a ground plane, with a uniformly sampled rotation around the normal direction of the ground plane. For each scene, we render 9 extra instances for evaluating novel view synthesis, and 9 for relighting. All scenes are shown in the supplement. 

\paragraph{Metrics.}
We evaluate the normal prediction accuracy by angle deviation from the ground truth, and the albedo prediction with scale-invariant metrics. We use the same similarity metrics as in \Cref{subsect:real} for novel-view synthesis, relighting, and albedo comparisons. We measure depth accuracy with the scale-invariant MSE error. All predictions are first applied with ground truth masks before computing the metrics. 

\paragraph{Baselines.}
While each training image receives a global illumination, the lighting configuration for each instance in the training image varies locally due to difference in poses. Therefore, we compare our model with two inverse rendering methods that handle varying light sources in training images, namely Neural-PIL~\cite{boss2021neuralpil} and NeRD~\cite{boss2021nerd}. 
Both methods jointly optimize for the shape, BRDF, and illumination from multi-view images with a NeRF~\cite{mildenhall2020nerf}-based architecture. Neural-PIL additionally proposes to use a neural network as a proxy for the illumination integration procedure in forward rendering.  
We also compare with NeRF and GNeRF, which only perform 3D reconstruction. 

\paragraph{Implementation Details.}
We train our method for 300k iterations for all scenes. 
Baseline methods are designed to train on a multi-view dataset, and we convert each scene into a multi-view dataset similar to \cref{subsect:real}. 
Neural-PIL, NeRD, and NeRF assume constant, ground-truth camera intrinsics across all multi-view images in the dataset. Since cropping around instances with different locations in the scene results in inconsistent intrinsic parameters, we re-render each instance in the scene so that it is re-centered at the origin, with its rotation remaining the same. For completeness, we also report the results of training these methods without the re-centering operation in the supplement.

\paragraph{Results.}
As shown in \Cref{tbl:synthetic}, our method achieves a significantly lower error in normal and appearance reconstruction compared to GNeRF, evaluated on held-out test views. The quality of both reconstruction and intrinsics decomposition is significantly better than Neural-PIL and NeRD, both requiring ground-truth camera poses, and is comparable with NeRF, which also requires ground-truth poses and does not perform intrinsic decomposition. 
The qualitative results for two exemplar scenes are shown in \Cref{fig:synthetic}. Full qualitative comparisons are deferred to the supplement. 

\section{Conclusion}
We have proposed a method that recovers the object intrinsics---the distributions of geometry, texture,
and material, 
separated from extrinsics such as poses and lighting, from a single image containing multiple instances of the same object type with masks. We have developed a neural representation to model such intrinsics and an adversarial framework for training. The proposed method successfully recovers object intrinsics on various objects from Internet images, enabling many applications including shape and image generation, novel view synthesis, and relighting.

\paragraph{Limitations.} The proposed method assumes that multiple similar, non-occluded instances are present in the input image. We leave modeling more cluttered scenes to future work. 
We also approximate the light source with a single directional light. While this simple parameterization achieves reasonable results for input images with more complex illumination effects, the model is not able to model shadows and inter-reflection and tends to bake them into the albedo. Rows 3--4 of \Cref{fig:wild} are examples of such failure cases.

\paragraph{Acknowledgments.}
We thank Angjoo Kanazawa, Josh Tenenbaum, Ruocheng Wang, Kai Zhang, Yiming Dou, and Koven Yu for their feedback. This work is in part supported by the Stanford Institute for Human-Centered AI (HAI), NSF CCRI \#2120095, NSF RI \#2211258, ONR MURI N00014-22-1-2740, AFOSR YIP FA9550-23-1-0127, Amazon, Bosch, Ford, Google, and Samsung.

{\small
\bibliographystyle{ieee_fullname}
\bibliography{reference}

\begin{thebibliography}{10}\itemsep=-1pt

\bibitem{blenderbun}
3{D} model bun.
\newblock \url{https://www.blendswap.com/blend/18213}.

\bibitem{blenderpeanut}
3{D} model peanut.
\newblock \url{https://www.blendswap.com/blend/17808}.

\bibitem{barron2012color}
Jonathan~T Barron and Jitendra Malik.
\newblock Color constancy, intrinsic images, and shape estimation.
\newblock In {\em ECCV}, 2012.

\bibitem{BarronTPAMI2015}
Jonathan~T. Barron and Jitendra Malik.
\newblock {Shape, Illumination, and Reflectance from Shading}.
\newblock {\em IEEE TPAMI}, 2015.

\bibitem{bolelli2019spaghetti}
Federico Bolelli, Stefano Allegretti, Lorenzo Baraldi, and Costantino Grana.
\newblock Spaghetti labeling: Directed acyclic graphs for block-based connected
  components labeling.
\newblock {\em IEEE TIP}, 2019.

\bibitem{boss2021nerd}
Mark Boss, Raphael Braun, Varun Jampani, Jonathan~T Barron, Ce Liu, and Hendrik
  Lensch.
\newblock Nerd: Neural reflectance decomposition from image collections.
\newblock In {\em ICCV}, 2021.

\bibitem{boss2021neuralpil}
Mark Boss, Varun Jampani, Raphael Braun, Ce Liu, Jonathan~T. Barron, and
  Hendrik~P.A. Lensch.
\newblock Neural-pil: Neural pre-integrated lighting for reflectance
  decomposition.
\newblock In {\em NeurIPS}, 2021.

\bibitem{piGAN2021}
Eric Chan, Marco Monteiro, Petr Kellnhofer, Jiajun Wu, and Gordon Wetzstein.
\newblock pi-gan: Periodic implicit generative adversarial networks for
  3d-aware image synthesis.
\newblock In {\em CVPR}, 2021.

\bibitem{chan2022eg}
Eric~R. Chan, Connor~Z. Lin, Matthew~A. Chan, Koki Nagano, Boxiao Pan,
  Shalini~De Mello, Orazio Gallo, Leonidas Guibas, Jonathan Tremblay, Sameh
  Khamis, Tero Karras, and Gordon Wetzstein.
\newblock Efficient geometry-aware {3D} generative adversarial networks.
\newblock In {\em CVPR}, 2022.

\bibitem{deng09imagenet}
Jia Deng, Wei Dong, Richard Socher, Li-Jia Li, Kai Li, and Li Fei-Fei.
\newblock Imagenet: A large-scale hierarchical image database.
\newblock In {\em CVPR}, 2009.

\bibitem{du2021nerflow}
Yilun Du, Yinan Zhang, Hong-Xing Yu, Joshua~B. Tenenbaum, and Jiajun Wu.
\newblock Neural radiance flow for 4d view synthesis and video processing.
\newblock In {\em ICCV}, 2021.

\bibitem{eigen2014depth}
David Eigen, Christian Puhrsch, and Rob Fergus.
\newblock Depth map prediction from a single image using a multi-scale deep
  network.
\newblock {\em NeurIPS}, 2014.

\bibitem{goodfellow2014generative}
Ian Goodfellow, Jean Pouget-Abadie, Mehdi Mirza, Bing Xu, David Warde-Farley,
  Sherjil Ozair, Aaron Courville, and Yoshua Bengio.
\newblock Generative adversarial nets.
\newblock In {\em NeurIPS}, 2014.

\bibitem{granot2022drop}
Niv Granot, Ben Feinstein, Assaf Shocher, Shai Bagon, and Michal Irani.
\newblock Drop the gan: In defense of patches nearest neighbors as single image
  generative models.
\newblock In {\em CVPR}, 2022.

\bibitem{gropp2020implicit}
Amos Gropp, Lior Yariv, Niv Haim, Matan Atzmon, and Yaron Lipman.
\newblock Implicit geometric regularization for learning shapes.
\newblock {\em ICML}, 2020.

\bibitem{haim2022diverse}
Niv Haim, Ben Feinstein, Niv Granot, Assaf Shocher, Shai Bagon, Tali Dekel, and
  Michal Irani.
\newblock Diverse generation from a single video made possible.
\newblock In {\em ECCV}, 2022.

\bibitem{heusel2017gans}
Martin Heusel, Hubert Ramsauer, Thomas Unterthiner, Bernhard Nessler, and Sepp
  Hochreiter.
\newblock Gans trained by a two time-scale update rule converge to a local nash
  equilibrium.
\newblock In {\em NeurIPS}, 2017.

\bibitem{karnewar20223ingan}
Animesh Karnewar, Tobias Ritschel, Oliver Wang, and Niloy Mitra.
\newblock {3inGAN}: Learning a {3D} generative model from images of a
  self-similar scene.
\newblock In {\em 3DV}, 2022.

\bibitem{Karras2020ada}
Tero Karras, Miika Aittala, Janne Hellsten, Samuli Laine, Jaakko Lehtinen, and
  Timo Aila.
\newblock Training generative adversarial networks with limited data.
\newblock In {\em NeurIPS}, 2020.

\bibitem{karras2019style}
Tero Karras, Samuli Laine, and Timo Aila.
\newblock A style-based generator architecture for generative adversarial
  networks.
\newblock In {\em CVPR}, 2019.

\bibitem{kingma2014adam}
Diederik~P Kingma and Jimmy Ba.
\newblock Adam: A method for stochastic optimization.
\newblock {\em ICLR}, 2015.

\bibitem{kulikov2022sinddm}
Vladimir Kulikov, Shahar Yadin, Matan Kleiner, and Tomer Michaeli.
\newblock Sinddm: A single image denoising diffusion model.
\newblock {\em arXiv preprint arXiv:2211.16582}, 2022.

\bibitem{laffont2012coherent}
Pierre-Yves Laffont, Adrien Bousseau, Sylvain Paris, Fr{\'e}do Durand, and
  George Drettakis.
\newblock {Coherent Intrinsic Images from Photo Collections}.
\newblock {\em ACM TOG}, 2012.

\bibitem{liu2020unsupervised}
Yunfei Liu, Yu Li, Shaodi You, and Feng Lu.
\newblock Unsupervised learning for intrinsic image decomposition from a single
  image.
\newblock In {\em CVPR}, 2020.

\bibitem{meng2021gnerf}
Quan Meng, Anpei Chen, Haimin Luo, Minye Wu, Hao Su, Lan Xu, Xuming He, and
  Jingyi Yu.
\newblock Gnerf: Gan-based neural radiance field without posed camera.
\newblock In {\em ICCV}, 2021.

\bibitem{mildenhall2020nerf}
Ben Mildenhall, Pratul~P Srinivasan, Matthew Tancik, Jonathan~T Barron, Ravi
  Ramamoorthi, and Ren Ng.
\newblock Nerf: Representing scenes as neural radiance fields for view
  synthesis.
\newblock In {\em ECCV}, 2020.

\bibitem{nguyen2019hologan}
Thu Nguyen-Phuoc, Chuan Li, Lucas Theis, Christian Richardt, and Yong-Liang
  Yang.
\newblock Hologan: Unsupervised learning of 3d representations from natural
  images.
\newblock In {\em ICCV}, 2019.

\bibitem{niemeyer2021giraffe}
Michael Niemeyer and Andreas Geiger.
\newblock Giraffe: Representing scenes as compositional generative neural
  feature fields.
\newblock In {\em CVPR}, 2021.

\bibitem{niemeyer2020differentiable}
Michael Niemeyer, Lars Mescheder, Michael Oechsle, and Andreas Geiger.
\newblock Differentiable volumetric rendering: Learning implicit 3d
  representations without 3d supervision.
\newblock In {\em CVPR}, 2020.

\bibitem{orel2021stylesdf}
Roy Or-El, Xuan Luo, Mengyi Shan, Eli Shechtman, Jeong~Joon Park, and Ira
  Kemelmacher-Shlizerman.
\newblock Style{SDF}: {H}igh-{R}esolution {3D}-{C}onsistent {I}mage and
  {G}eometry {G}eneration.
\newblock In {\em CVPR}, 2022.

\bibitem{park2021nerfies}
Keunhong Park, Utkarsh Sinha, Jonathan~T. Barron, Sofien Bouaziz, Dan~B
  Goldman, Steven~M. Seitz, and Ricardo Martin-Brualla.
\newblock Nerfies: Deformable neural radiance fields.
\newblock {\em ICCV}, 2021.

\bibitem{pumarola2020d}
Albert Pumarola, Enric Corona, Gerard Pons-Moll, and Francesc Moreno-Noguer.
\newblock {D-NeRF: Neural Radiance Fields for Dynamic Scenes}.
\newblock In {\em CVPR}, 2020.

\bibitem{qin2020u2}
Xuebin Qin, Zichen Zhang, Chenyang Huang, Masood Dehghan, Osmar~R Zaiane, and
  Martin Jagersand.
\newblock U2-net: Going deeper with nested u-structure for salient object
  detection.
\newblock {\em Pattern Recognition}, 2020.

\bibitem{rombach2021highresolution}
Robin Rombach, Andreas Blattmann, Dominik Lorenz, Patrick Esser, and Björn
  Ommer.
\newblock High-resolution image synthesis with latent diffusion models.
\newblock {\em arXiv preprint arXiv:2112.10752}, 2021.

\bibitem{schoenberger2016sfm}
Johannes~Lutz Sch\"{o}nberger and Jan-Michael Frahm.
\newblock Structure-from-motion revisited.
\newblock In {\em CVPR}, 2016.

\bibitem{shaham2019singan}
Tamar~Rott Shaham, Tali Dekel, and Tomer Michaeli.
\newblock Singan: Learning a generative model from a single natural image.
\newblock In {\em ICCV}, 2019.

\bibitem{shocher2019ingan}
Assaf Shocher, Shai Bagon, Phillip Isola, and Michal Irani.
\newblock Ingan: Capturing and retargeting the" dna" of a natural image.
\newblock In {\em ICCV}, 2019.

\bibitem{sitzmann2019siren}
Vincent Sitzmann, Julien~N.P. Martel, Alexander~W. Bergman, David~B. Lindell,
  and Gordon Wetzstein.
\newblock Implicit neural representations with periodic activation functions.
\newblock In {\em NeurIPS}, 2020.

\bibitem{verbin2022ref}
Dor Verbin, Peter Hedman, Ben Mildenhall, Todd Zickler, Jonathan~T Barron, and
  Pratul~P Srinivasan.
\newblock Ref-nerf: Structured view-dependent appearance for neural radiance
  fields.
\newblock In {\em CVPR}, 2022.

\bibitem{wang2021neus}
Peng Wang, Lingjie Liu, Yuan Liu, Christian Theobalt, Taku Komura, and Wenping
  Wang.
\newblock Neus: Learning neural implicit surfaces by volume rendering for
  multi-view reconstruction.
\newblock In {\em NeurIPS}, 2021.

\bibitem{wang2022sindiffusion}
Weilun Wang, Jianmin Bao, Wengang Zhou, Dongdong Chen, Dong Chen, Lu Yuan, and
  Houqiang Li.
\newblock Sindiffusion: Learning a diffusion model from a single natural image.
\newblock {\em arXiv preprint arXiv:2211.12445}, 2022.

\bibitem{wang2022singrav}
Yujie Wang, Xuelin Chen, and Baoquan Chen.
\newblock Singrav: Learning a generative radiance volume from a single natural
  scene.
\newblock {\em arXiv preprint arXiv:2210.01202}, 2022.

\bibitem{wang2004image}
Zhou Wang, Alan~C Bovik, Hamid~R Sheikh, and Eero~P Simoncelli.
\newblock Image quality assessment: from error visibility to structural
  similarity.
\newblock {\em IEEE TIP}, 2004.

\bibitem{wimbauer2022derender}
Felix Wimbauer, Shangzhe Wu, and Christian Rupprecht.
\newblock De-rendering 3d objects in the wild.
\newblock In {\em CVPR}, 2022.

\bibitem{wu2021derender}
Shangzhe Wu, Ameesh Makadia, Jiajun Wu, Noah Snavely, Richard Tucker, and
  Angjoo Kanazawa.
\newblock De-rendering the world's revolutionary artefacts.
\newblock In {\em CVPR}, 2021.

\bibitem{yariv2020multiview}
Lior Yariv, Yoni Kasten, Dror Moran, Meirav Galun, Matan Atzmon, Ronen Basri,
  and Yaron Lipman.
\newblock {Multiview Neural Surface Reconstruction by Disentangling Geometry
  and Appearance}.
\newblock In {\em NeurIPS}, 2020.

\bibitem{YuSelfRelight20}
Ye Yu, Abhimetra Meka, Mohamed Elgharib, Hans-Peter Seidel, Christian Theobalt,
  and Will Smith.
\newblock {Self-supervised Outdoor Scene Relighting}.
\newblock In {\em ECCV}, 2020.

\bibitem{zhang2021physg}
Kai Zhang, Fujun Luan, Qianqian Wang, Kavita Bala, and Noah Snavely.
\newblock Physg: Inverse rendering with spherical gaussians for physics-based
  material editing and relighting.
\newblock In {\em CVPR}, 2021.

\bibitem{zhang2018unreasonable}
Richard Zhang, Phillip Isola, Alexei~A. Efros, Eli Shechtman, and Oliver Wang.
\newblock The unreasonable effectiveness of deep networks as a perceptual
  metric.
\newblock In {\em CVPR}, 2018.

\bibitem{zhang2021nerfactor}
Xiuming Zhang, Pratul~P Srinivasan, Boyang Deng, Paul Debevec, William~T
  Freeman, and Jonathan~T Barron.
\newblock Nerfactor: Neural factorization of shape and reflectance under an
  unknown illumination.
\newblock In {\em SIGGRAPH Asia}, 2021.

\bibitem{zhou2019continuity}
Yi Zhou, Connelly Barnes, Jingwan Lu, Jimei Yang, and Hao Li.
\newblock On the continuity of rotation representations in neural networks.
\newblock In {\em CVPR}, 2019.

\end{thebibliography}
}

\end{document}